\documentclass[letterpaper, 10 pt, conference]{ieeeconf}  

\IEEEoverridecommandlockouts                              

\usepackage{graphicx, tipa}
\usepackage{amsmath,amssymb,amsfonts}
\usepackage[caption=false]{subfig}
\usepackage[ruled,linesnumbered, noend]{algorithm2e}
\usepackage{dirtytalk}
\usepackage{hyperref}
\usepackage{gensymb}
\usepackage{tabularx}
\usepackage{multirow}
\usepackage[short]{optidef}
\usepackage{booktabs}
\usepackage{siunitx}
\usepackage{titlesec}
\newcounter{cases}
\newcounter{subcases}[cases]
\newenvironment{mycase}
{
    \setcounter{cases}{0}
    \setcounter{subcases}{0}
    \newcommand{\case}
    {
        \par\indent\stepcounter{cases}\textbf{Case \thecases.}
    }
    
}
{
    \par
}
\renewcommand*\thecases{\arabic{cases}}

\newcommand{\arc}[1]{{%
  \setbox9=\hbox{#1}%
  \ooalign{\resizebox{\wd9}{\height}{\texttoptiebar{\phantom{A}}}\cr#1}}}

\makeatletter
\newcommand{\removelatexerror}{\let\@latex@error\@gobble}
\makeatother
\SetKwComment{Comment}{$\triangleright$\ }{}

\newcommand\Tstrut{\rule{0pt}{2.0ex}}         

\title{\LARGE \bf
Vision-aided UAV Navigation and Dynamic Obstacle Avoidance using Gradient-based B-spline Trajectory Optimization}

\author{Zhefan Xu, Yumeng Xiu, Xiaoyang Zhan, Baihan Chen, and Kenji Shimada 
\thanks{Zhefan Xu, Yumeng Xiu, Xiaoyang Zhan, Baihan Chen, and Kenji Shimada are with the Department of Mechanical Engineering, Carnegie Mellon University, 5000 Forbes Ave, Pittsburgh, PA, 15213, USA.
        {\tt\small zhefanx@andrew.cmu.edu}}%
}

\begin{document}

\maketitle
\thispagestyle{empty}
\pagestyle{empty}

\begin{abstract}
Navigating dynamic environments requires the robot to generate collision-free trajectories and actively avoid moving obstacles. Most previous works designed path planning algorithms based on one single map representation, such as the geometric, occupancy, or ESDF map. Although they have shown success in static environments, due to the limitation of map representation, those methods cannot reliably handle static and dynamic obstacles simultaneously. To address the problem, this paper proposes a gradient-based B-spline trajectory optimization algorithm utilizing the robot's onboard vision. The depth vision enables the robot to track and represent dynamic objects geometrically based on the voxel map. The proposed optimization first adopts the circle-based guide-point algorithm to approximate the costs and gradients for avoiding static obstacles. Then, with the vision-detected moving objects, our receding-horizon distance field is simultaneously used to prevent dynamic collisions. Finally, the iterative re-guide strategy is applied to generate the collision-free trajectory. The simulation and physical experiments prove that our method can run in real-time to navigate dynamic environments safely. Our software is available on GitHub\footnote{\url{https://github.com/Zhefan-Xu/CERLAB-UAV-Autonomy}} as an open-source package.
\end{abstract}

\section{Introduction}
Light-weight autonomous UAVs are massively deployed in various industrial applications, such as inspection, exploration, and search and rescue. The environments of those applications are usually highly complex and dynamic, involving human workers, static structures, robots, and vehicles. As one of the most fundamental components of robot autonomy, the safe trajectory planning algorithm becomes essential to let UAVs deal with complex environment structures while actively sensing and avoiding dynamic obstacles.

Safe navigation in dynamic environments mainly involves three challenges. Firstly, the robot must track and represent static and dynamic obstacles simultaneously. Recent popular planning methods \cite{fastplanner}\cite{egoplanner} apply the voxel-based map including the occupancy map and ESDF map \cite{voxblox}\cite{fiesta} as obstacle representation. Although those mapping algorithms can deal with arbitrarily complex static environments, they can hardly distinguish and capture dynamic obstacles, leading to the limited performance of the mentioned planners. Secondly, the planner should be able to generate trajectories with complicated static structures. Some vision-based algorithms \cite{reactiveUV}\cite{CCMPC}\cite{visionCCMPC} represent obstacles geometrically using the bounding boxes or ellipsoids. Those methods can safely avoid dynamic obstacles but might fail when environmental structures become complicated. Finally, due to the unpredictable and quickly changing environments, high-frequency real-time planning is necessary to prevent dynamic collisions, which further adds burden to the limited onboard computations. 

\begin{figure}[t] 
    \vspace{0.2cm}
    \centering
    \includegraphics[scale=0.6]{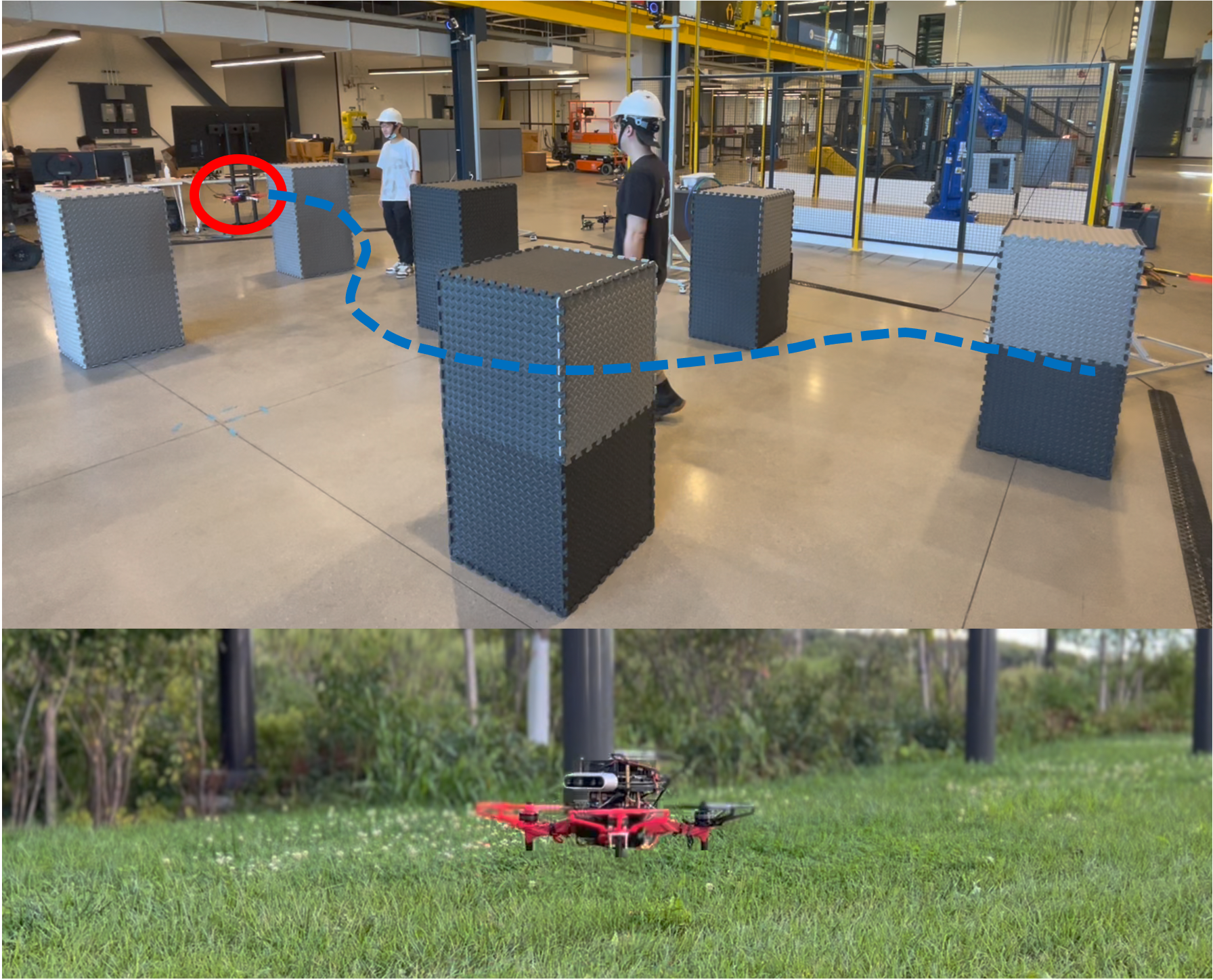}
    \caption{The UAV navigating with obstacles using the proposed algorithm. The upper presents the physical autonomous flight in a dynamic environment. The bottom shows our customized UAV with onboard sensors.}
    \label{intro_figure}
\end{figure}

To solve these issues, this paper proposes the \textbf{Vi}sion-aided \textbf{G}radient-based B-spline Trajectory \textbf{O}ptimization (\textbf{ViGO}) algorithm.  The algorithm utilizes our vision-aided 3D dynamic map, enabling tracking dynamic obstacles and representing complex static environments simultaneously. The proposed circle-based guide-point algorithm approximates the costs and gradients of static collision to improve optimization speed. With future predictions, the receding horizon distance field is applied to prevent collisions considering dynamic obstacles. Finally, the iterative re-guide strategy is used to real-time generate collision-free trajectories. Our customized UAV and the dynamic environment navigation example are shown in Fig. \ref{intro_figure}, and the main contributions of this work are: 

\begin{itemize}
    \item \textbf{Vision-aided 3D Dynamic Map:} This algorithm utilizes the depth image with the occupancy voxel map to track 3D dynamic obstacles and adopts this 3D dynamic map to perform the trajectory optimization.
    \item \textbf{Circle-based Guide-Point Algorithm:} We propose the guide-point algorithm to approximate the costs and gradients of static collisions for trajectory optimization.
    \item \textbf{Receding Horizon Distance Field:} The proposed receding horizon distance field utilizes the obstacles' future predictions to estimate the collision costs and gradients for preventing dynamic collisions.
\end{itemize}

\section{Related Work}
Recent years have seen many works of UAV navigation in dynamic environments. There are mainly two categories of methods based on their map representations: the voxel map-based method and the geometric map-based method. With the advantage of representing arbitrary complex 3D objects, the voxel maps such as \cite{voxblox}\cite{fiesta}\cite{octomap} are widely used in trajectory planning. Due to the differential flatness property \cite{minsnap} of the quadcopter, \cite{polynomial} generates the trajectory using the minimum snap optimization, iteratively adding intermediate waypoints to prevent collisions. Inspired by \cite{chomp}\cite{stomp}, \cite{fastplanner} formalizes the unconstrained optimization utilizing the distance information from the ESDF map. Later, \cite{egoplanner} applies the collision force to avoid obstacles locally and reduce the ESDF update computation. In \cite{DEP}, incremental sampling is proposed to reduce the computation time for dynamic environment exploration. Although the above methods prove high computational efficiency and success in aggressive flights, their map cannot capture and distinguish the dynamic obstacle well, leading to a limited performance in highly dynamic environments.

Unlike the voxel map, the geometric map usually represents each obstacle as a single bounding box, sphere, or ellipsoid. Based on the analytical form of those geometries, early approaches propose the artificial potential field \cite{APF} and velocity obstacles \cite{orca} to generate simple control commands to prevent collisions. Recently, the model predictive control-based (MPC) methods \cite{robust17}\cite{chance11}\cite{CCMPC}\cite{chance20}\cite{guo2022obstacle} have become popular for obstacle avoidance while considering robot dynamics. In \cite{robust17}, the distance cost is applied for each obstacle for the trajectory penalty. \cite{chance11} adopts the chance-constrained formulation and reduces computation by disjunctive programming to consider the uncertainty.
Similarly, \cite{CCMPC}\cite{chance20} approximates the chance constraint by linearization and achieves real-time obstacle avoidance. Those mentioned methods assume the obstacles can always be represented using geometric shapes and rely on vision-based obstacle detection \cite{chance20}\cite{reactiveUV}. However, when the static environmental structure becomes complex, those geometric shapes might become over conservative or even unable to model static obstacles, resulting in collisions and suboptimal performance.

A few methods also distinguish static and dynamic obstacles for trajectory planning. A two-layer planner scheme is proposed in \cite{DPMPC} to deal with static and dynamic obstacles simultaneously. With the dual-structure particle-based map, \cite{dsp} adopts the sampling-based planner to evaluate the risk of trajectories. However, their solutions can be suboptimal since they only sample from the predefined motion primitives. In \cite{eth_detection}, learning-based detection is applied to track the dynamic obstacles with the occupancy map. \cite{visionZJU} applies point cloud-based detection with the voxel map and uses linear prediction of obstacle state to achieve safe navigation. However, their method of fully trusting linear predicted future states might be over-conservative and even fail to find the solution in some narrow environments. Unlike the previous methods, our approach utilizes the vision-aided 3D voxel map to track dynamic obstacles and applies the gradient-based trajectory optimization considering the dynamic obstacles' future states in a receding horizon manner.

\section{Methodology} \label{method}
This section presents each component of our proposed trajectory optimization framework. In Sec. \ref{mapping}, the vision-aided 3D dynamic map system is introduced. Then, the unconstrained B-spline trajectory optimization is defined in Sec. \ref{optimization}. The static collision cost using our circle-based guide point algorithm is discussed for static obstacle avoidance in Sec. \ref{static_cost}. Then, we formalize the receding horizon distance field for calculating the dynamic obstacle cost in Sec. \ref{dynamic_cost}. Finally, Sec. \ref{reguide_optimization} introduces the iterative re-guide optimization to generate the collision-free trajectory.

\subsection{Vision-aided 3D Dynamic Mapping} \label{mapping}
\begin{figure}[t] 
    \vspace{0.1cm}
    \centering
    \includegraphics[scale=0.50]{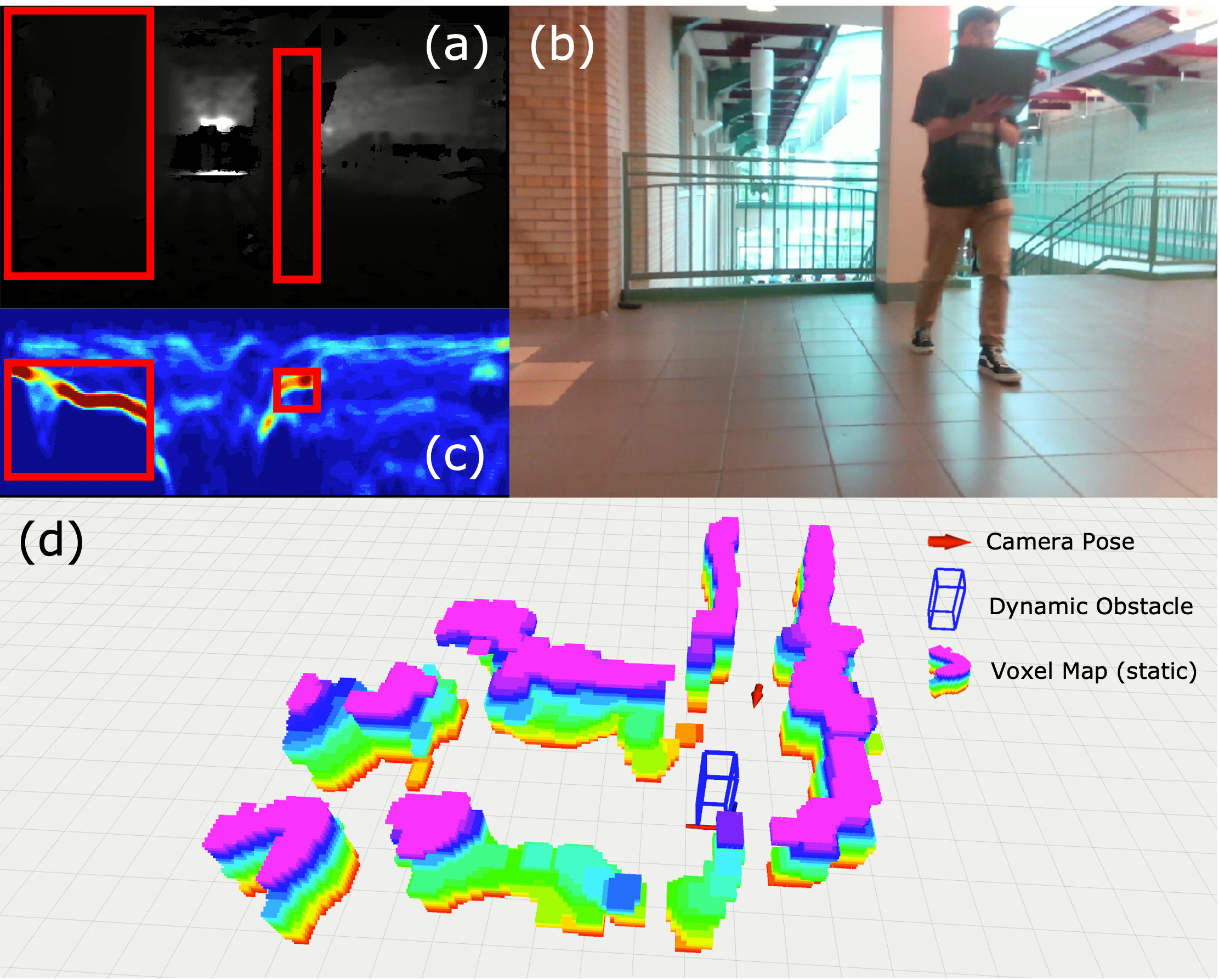}
    \caption{Illustration of the 3D dynamic map. (a) Depth image with detected objects. (b) RGB camera view. (c) U-depth map generated by the depth image with the detected results. (d) Visualization of the dynamic map.}
    \label{dynamic_map}
\end{figure}

This part introduces a lightweight mapping method to handle static and dynamic obstacles simultaneously.The dynamic map method has become popular in recent years to track dynamic voxels \cite{dynamicMap18}\cite{dynamicMap21_1}\cite{dynamicMap21_2}. Unlike those methods, this method is based on the 3D occupancy map and utilizes the depth image to perform dynamic obstacle tracking. There are mainly four steps to update the map and track dynamic obstacles: region proposal detection, map-depth fusion, dynamic obstacle filtering, and dynamic region cleaning.

The proposed method first applies the depth image-based detection to find the possible obstacle regions as the proposals. Given the input depth image, we first generate the u-depth map shown in Fig. \ref{dynamic_map}c, as described in \cite{reactiveUV}\cite{chance20}. The u-depth map can be interpreted as the top view of the objects, so the lengths and widths of potential objects can be detected as bounding boxes by line grouping shown in Fig. \ref{dynamic_map}c. Then, with obstacles' widths, the bounding boxes of objects can be found in the original depth image by depth continuity checking. After performing the coordinate transformation, we can obtain 3D bounding boxes of potential objects in the world frame. Note that the obtained bounding boxes only preserve the objects' rough positions and sizes and might include static obstacles. To refine the positions and sizes of obstacles as the second step, we enlarge those detected bounding boxes as the region proposals by an inflating factor and check the occupancy information in the voxel map, which stores discrete but more accurate information. Through this region proposal search, we can get refined results.

With the refined bounding boxes, the third step is to filter out static obstacles. We first apply the Kalman filter to track and determine the velocity of each detected object. We can remove most static obstacles by using the minimum velocity threshold criteria. However, the detection noises make some static obstacles shake back and forth slightly, leading to some velocity errors. To solve this issue, we apply our continuity filter to remove objects with jerky motions using the history of object velocities. Finally, since the static occupancy map might still keep the occupied voxel of dynamic obstacles, we apply the histories of dynamic obstacles to clean those regions. The final map output is shown in Fig. \ref{dynamic_map}d.


\subsection{B-spline Trajectory Optimization} \label{optimization}
B-spline curves are widely used in trajectory optimization \cite{fastplanner}\cite{egoplanner} due to its local curve control ability, convex hull property, etc. A B-spline curve with order $k$ is composed of several $k-1$ degree polynomial basis functions with corresponding control points defined over a knot vector. Given the global trajectory or the goal position, we can parameterize the trajectory into a set of control points:
\begin{equation}
\mathcal{\hat{S}} = \{\textbf{P}_{1}, \textbf{P}_{2}, \textbf{P}_{3}, ..., \textbf{P}_{N-1}, \textbf{P}_{N}\},  \  \   \textbf{P}_{i} \in \mathcal{R}^\text{3},
\end{equation}
where the first and last $k-1$ control points are the position of start and goal, respectively. The optimization variables are the set $\mathcal{S}$ of the intermediate $N-2(k-1)$ control points. 

The optimization follows the gradient-based unconstrained formulation. With the variable set $\mathcal{S}$, the objective cost function is defined as:
\begin{equation} \label{objective_function}
\begin{split}
    \text{C}_{\text{total}}(\mathcal{S}) = \alpha_{\text{control}} \cdot \text{C}_{\text{control}} + \alpha_{\text{smooth}} \cdot \text{C}_{\text{smooth}} \\ + \alpha_{\text{static}} \cdot \text{C}_{\text{static}} + \alpha_{\text{dynamic}} \cdot \text{C}_{\text{dynamic}},  
\end{split}
\end{equation}
which is a weighted combination of control limit, trajectory smoothness, static collision, and dynamic obstacle costs. The static collision cost and dynamic obstacle cost will be discussed in detail in Sec. \ref{static_cost} and \ref{dynamic_cost}.

The control limit cost forces the trajectory to have feasible velocities and accelerations. The derivative of the B-spline curve can be represented by another B-spline, so the control points $\textbf{V}_{i}$ and $\textbf{A}_{i}$ for velocity and acceleration are:
\begin{equation}
    \textbf{V}_{i} = \frac{\textbf{P}_{i+1} - \textbf{P}_{i}}{\delta t}, \ \textbf{A}_{i} = \frac{\textbf{V}_{i+1} - \textbf{V}_{i}}{\delta t}, 
\end{equation}
where $\delta t$ is the time step. So, given the maximum velocity $\textbf{v}_{\text{max}}$ and acceleration $\textbf{a}_{\text{max}}$, the cost function is defined as:
\begin{equation}
    \text{C}_{\text{control}} = \sum_{i} {\frac{||\textbf{V}_{i} - \textbf{v}_{\text{max}}||_{2}^{2}}{\lambda_{\text{vel}}} + \frac{||\textbf{A}_{i} - \textbf{a}_{\text{max}}||_{2}^{2}}{\lambda_{\text{acc}}}}, \label{control_cost}
\end{equation}
in which the L2 norm is applied for the control limit penalty with the unit normalization factor $\lambda$ when the velocity and acceleration exceed the limits.

The smoothness cost is applied to prevent the jerky trajectory. Following the same manner, we can get the control points $\textbf{J}_{i}$ for the jerk, the derivative of acceleration, and apply the following cost function:
\begin{equation}
    \text{C}_{\text{smooth}} = \sum_{i} {||\textbf{J}_{i}||_{2}^{2}}, \ \  \textbf{J}_{i} = \frac{\textbf{A}_{i+1} - \textbf{A}_{i}}{\delta t} 
\end{equation}

\subsection{Static Collision Cost} \label{static_cost}
\begin{figure}[t] 
    \vspace{0.2cm}
    \centering
    \includegraphics[scale=0.48]{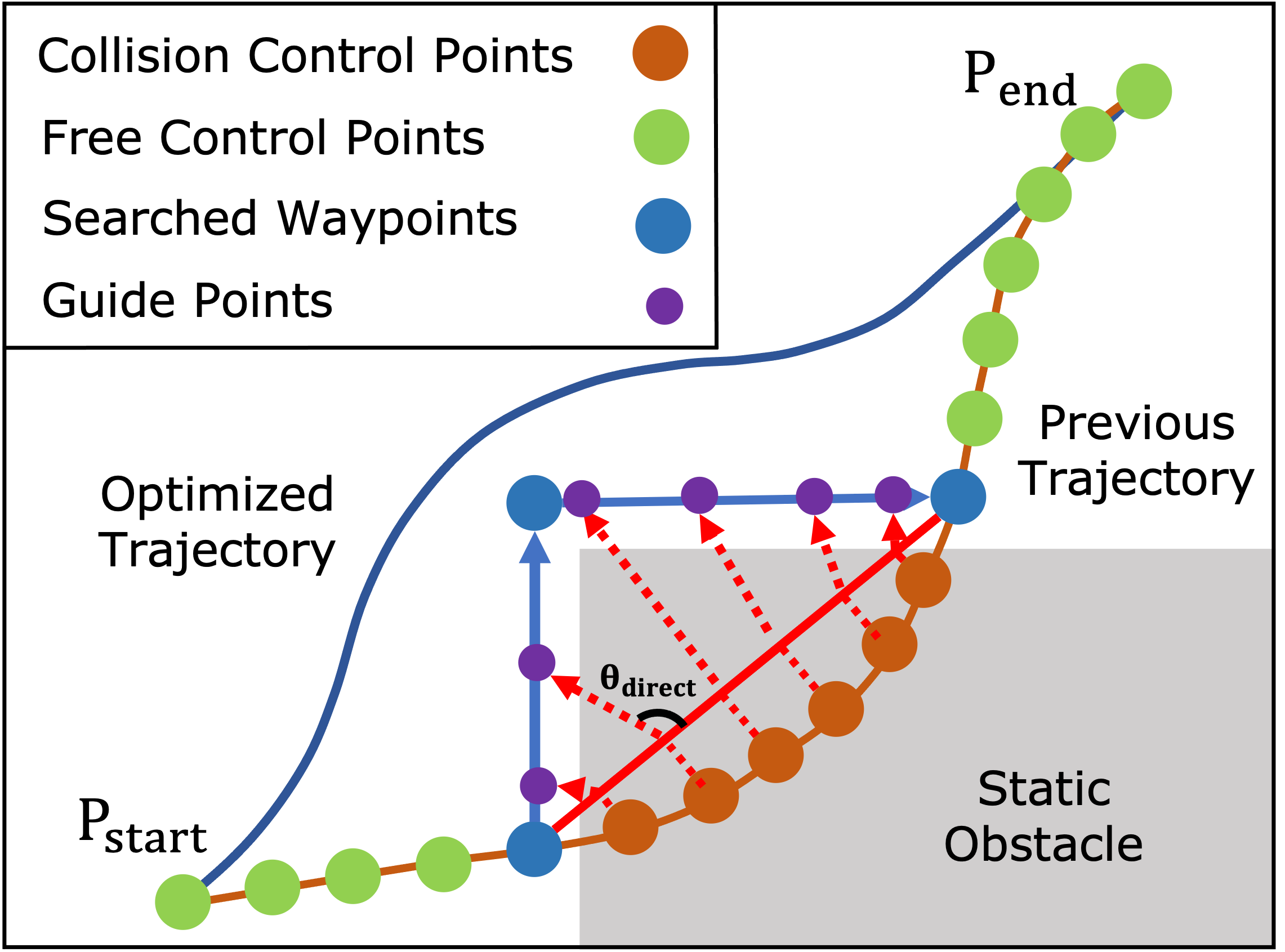}
    \caption{Illustration of circle-based guide-point assignment. For the given collision trajectory, we first find the collision control points and search the collision-free paths. The guide points shown as purple points are the intersections between circle-based raycasting and the searched paths.}
    \label{guide_point}
\end{figure}

\begin{algorithm}[t] \label{guide_point_algorithm}
\caption{Circle-based Guide-Point Algorithm} 
\SetAlgoNoLine%
$\mathcal{C}_{\text{guide}} \gets  \emptyset$ \Comment*[r]{guide-point set}
$\mathcal{\sigma}_{\text{traj}} \gets \text{collision trajectory}$\;
$\mathcal{S}_{\text{col}} \gets \textbf{findCollisionSet}(\mathcal{\sigma}_{\text{traj}})$\;
$\mathcal{P}_{\text{search}} \gets \textbf{pathSearch}(\mathcal{S}_{\text{col}}.\text{start}, \mathcal{S}_{\text{col}}.\text{end})$\;
$\mathcal{L} \gets \textbf{formLine}(\mathcal{P}_{\text{search}}.\text{start}, \mathcal{P}_{\text{search}}.\text{end})$\;
$n \gets 1;\  \mathcal{N} \gets  \mathcal{S}_{\text{col}}.\text{size}+1$\; 
\For{$\normalfont{\textbf{P}}_{\normalfont{\text{c}}}$ \normalfont{\textbf{in}} $\mathcal{S}_{\normalfont{\text{col}}}$}{
    $\textbf{P}_{\text{proj}} \gets \normalfont{\textbf{P}}_{\normalfont{\text{c}}}.\textbf{projectOnto}(\mathcal{L})$\; \label{project_point}
    $\mathcal{\theta}_{\text{direct}} \gets \frac{n}{\mathcal{N}} \cdot \pi$ \Comment*[r]{direction angle}
    $\textbf{P}_{\text{guide}} \gets \textbf{raycastOntoPath}(\textbf{P}_{\text{proj}}, \mathcal{\theta}_{\text{direct}}, \mathcal{P}_{\text{search}})$\;
    $\mathcal{C}_{\text{guide}}.\textbf{insert}(\textbf{P}_{\text{c}},\textbf{P}_{\text{guide}})$\;
    $n \gets n+1$
}
$\textbf{return} \ \mathcal{C}_{\text{guide}}$\;
\end{algorithm}

Since the static obstacles are represented by the occupancy voxels on the map, the collision costs and gradients for the above optimization cannot be directly obtained. So, we apply the proposed circle-based guide-point algorithm to estimate the costs and gradients of static collisions. 

The costs and gradients for the optimization variable, control points, can be estimated by the guide points. The procedure of finding guide points is illustrated in Fig. \ref{guide_point} and Alg. \ref{guide_point_algorithm}. Given a collision trajectory, the planner first finds the "collision set" $\mathcal{S}_{\text{col}}$, which consists of collision control points (Fig. \ref{guide_point} orange points). Inspired by \cite{egoplanner}, we apply the path-search algorithm, such as A* or Dijkstra, to locally find a collision-free path $\mathcal{P}_{\text{search}}$ bypassing the collision set. Then, we project each point in the collision set onto the connecting line between the start and goal position of the searched path (Alg. \ref{guide_point_algorithm} Line \ref{project_point}). With the projected point, we calculate the direction angles $\theta_{\text{direct}}$ ranging from $(0, \pi)$ and use the point-angle pair to cast a ray onto the searched path. The intersection point between the ray and the searched path is the guide point for that collision control point shown as purple points in Fig. \ref{guide_point}. The algorithm is circle-based because the direction angles sweep a semi-circle.

With the associated guide point and a predefined safety distance $\text{d}_{\text{safe}}$, the collision cost of collision control points can be calculated by:
\begin{equation}
    \text{C}_{\text{static}} = \sum_{i} \biggl(\textbf{max}\Bigl(\text{d}_{\text{safe}} - \textbf{signDist}(\textbf{P}_{\text{i}}, \textbf{P}^{\text{i}}_{\text{guide}}), 0\Bigl)\biggl)^3, \label{static_cost_function}
\end{equation}
where the signed distance defines the positive and negative distances as the control point outside and inside the obstacle, respectively. Eqn. \ref{static_cost_function} penalizes the control points whose signed distance is less than the safe distance using the cubic function. In this way, the gradient for \textbf{each collision control point} can be calculated with the chain rule:
\begin{equation}
    \nabla g_{i} = 3\Bigl(\text{d}_{\text{safe}} - \textbf{signDist}(\textbf{P}_{\text{i}}, \textbf{P}^{\text{i}}_{\text{guide}})\Bigl)^2 \cdot \ \frac{\hat{\textbf{P}}_{\text{i}} - \textbf{P}^{\text{i}}_{\text{guide}}}{||\hat{\textbf{{P}}}_{\text{i}} - \textbf{P}^{\text{i}}_{\text{guide}}||_{2}},
\end{equation}
where $\hat{\textbf{P}}_{\text{i}}$ is the initial collision control point, and the directions of the negative gradients are to push the collision control points to a free region, and Fig. \ref{guide_point} shows the optimized trajectory got a smooth turning to avoid the static obstacle by using the circle-based guide points.

\subsection{Dynamic Obstacle Cost} \label{dynamic_cost}
\begin{figure}[t] 
    \vspace{0.1cm}
    \centering
    \includegraphics[scale=0.42]{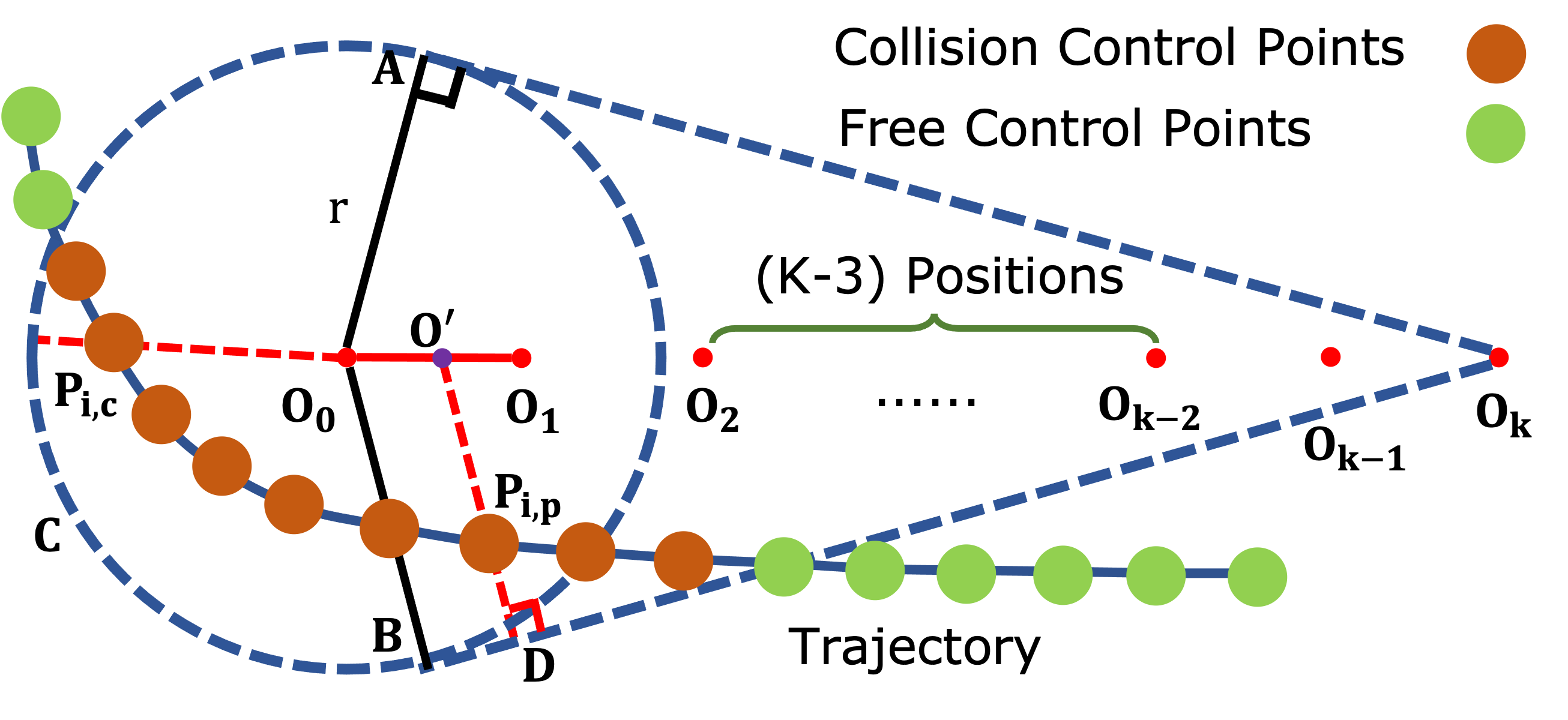}
    \caption{Illustration of receding horizon distance field. We linearly decrease the safety distance $\text{r}$ from the current obstacle position $\textbf{O}_{0}$ to the last predicted position $\textbf{O}_{k}$ in a receding horizon manner.}
    \label{distance_field}
\end{figure}
Unlike static obstacles, the states of dynamic obstacles are changing and uncertain, making it unreliable to use only the current captured information for trajectory optimization. So, this section introduces the receding horizon distance field for collision cost estimation, which evaluates the distance to the safe area of the given points based on future predictions.  

The illustration of the receding horizon distance field for dynamic obstacles is shown in Fig. \ref{distance_field}. The dynamic obstacles are represented as positions with estimated velocities and sizes. Given the obstacle's current position $\textbf{O}_{0}$ and velocity $\textbf{V}_{0}$, we apply the linear predictor to obtain the k steps' future positions $\{\textbf{O}_{1}, \textbf{O}_{2}, ..., \textbf{O}_\text{k}\}$. To construct the distance field, we draw a circle with the safe radius $\text{r}$ centered at $\textbf{O}_{0}$. Then, considering the unreliability of future predictions, we linearly decrease the safe radius $\text{r}$ to zero in a receding horizon manner, leading to a cone-shape collision region (blue dot lines). By avoiding this region, we can generate the trajectory considering future predictions while preventing over-conservative behaviors. So, for dynamic obstacle cost estimation, we first calculate their distances to the safe area considering the following two cases:
\begin{mycase}
    \case The control point $\textbf{P}_{\text{i,c}}$ is in the circular region enclosed by arc \arc{ACB}, line $\textbf{A}\textbf{O}_{0}$ and $\textbf{B}\textbf{O}_{0}$. The distance to safe area $\Delta \text{d}_{\text{i}}$ can be calculated by: 
    \begin{equation}
        \Delta \text{d}_{\text{i}} = \text{r} - ||\textbf{P}_{i,c} - \textbf{O}_{0}||_{2}, \label{case1}
    \end{equation}
    
    \case The control point $\textbf{P}_{\text{i,p}}$ is in the polygon region $\textbf{A}\textbf{O}_{0}\textbf{B}\textbf{O}_{k}$. We draw a line through $\textbf{P}_{\text{i,p}}$ which is perpendicular to the line $\textbf{B}\textbf{O}_{k}$ and intersects $\textbf{O}_{0}\textbf{O}_{k}$ at $\textbf{O}^{'}$. The distance to safe area $\Delta \text{d}_{i}$ can be calculated by: 
    \begin{equation}
        \Delta \text{d}_{\text{i}} = ||\textbf{D} - \textbf{O}^{'}||_{2} - ||\textbf{P}_{i,p} - \textbf{O}^{'}||_{2}. \label{case2}
    \end{equation}
\end{mycase}
\noindent With the distance to safe area $\Delta \text{d}_{\text{i}}$, we can finally calculate the costs for all the collision control points by:
\begin{equation}
    \text{C}_{\text{dynamic}} = \sum_{i} \Bigl(\textbf{max}(\Delta \text{d}_{\text{i}}, 0)\Bigl)^3, \label{dynamic_cost_function}
\end{equation}
and the gradient for each collision control point can be calculated using the chain rule with Eqns. \ref{case1} and \ref{case2}.

\subsection{Iterative Re-guide Optimization} \label{reguide_optimization}
\begin{algorithm}[t] \label{reguide_algorithm}
\caption{Iterative Re-guide Optimization} 
\SetAlgoNoLine%
 $\alpha_{\text{static}}, \alpha_{\text{dynamic}} \gets \alpha_{0}$ \Comment*[r]{Cost Weights}  \label{cost_weights}
 $\lambda \gets 1.5$ \Comment*[r]{user-defined inflate factor} \label{shrink_factor} \label{inflate_factor}
 $\mathcal{P} \gets \text{Optimization Solver}$\;
 $\mathcal{S} \gets \text{Control Points}$\;
 $\mathcal{T}_{\text{sc}} \gets true$ \Comment*[r]{static collision}
 $\mathcal{T}_{\text{dc}} \gets true$ \Comment*[r]{dynamic collision}
 
 \While{$\mathcal{T}_{\normalfont{\text{sc}}}$ \normalfont{\textbf{or}} $\mathcal{T}_{\textnormal{\text{dc}}}$}{ 
    $\text{reguide}, \mathcal{S}_{\text{reguide}} = \mathcal{P}.\textbf{isReguideRequired}(\mathcal{S})$\; \label{check_reguide}
    \If{$\normalfont{\text{reguide}}$ \normalfont{\textbf{or}} \normalfont{firstTime}}{
        $\mathcal{P}.\textbf{assignNewGuidePoints}(\mathcal{S}_{\text{reguide}})$ \label{assign_new_guide_point}
    }
    \ElseIf{\normalfont{\textbf{not}} $\normalfont{\text{reguide}}$ \normalfont{\textbf{and}} $\mathcal{T}_{\normalfont{\text{sc}}}$}{
        $\alpha_{\text{static}} \gets \lambda \cdot \alpha_{\text{static}}$ \label{inflate_static_weight}
    }
    \ElseIf{\normalfont{\textbf{not}} $\normalfont{\text{reguide}}$ \normalfont{\textbf{and}} $\mathcal{T}_{\normalfont{\text{dc}}}$}{
        $\alpha_{\text{dynamic}} \gets \lambda \cdot \alpha_{\text{dynamic}}$ \label{inflate_dynamic_weight}
    }
    $\mathcal{S} \gets \mathcal{P}.\textbf{updateCostAndSolve}()$\;
    $\mathcal{\sigma}_{\text{traj}} \gets \mathcal{P}.\textbf{evaluateTrajectory}(\mathcal{S})$\;
    $\mathcal{T}_{\text{sc}} \gets \mathcal{P}.\textbf{checkStaticCollision}(\mathcal{\sigma}_{\text{traj}})$\;
    $\mathcal{T}_{\text{dc}} \gets \mathcal{P}.\textbf{checkDynamicCollision}(\mathcal{\sigma}_{\text{traj}})$\;
 } 
$\textbf{return} \ \mathcal{\sigma}_{\text{traj}}$\;
\end{algorithm}
The proposed optimization problem is unconstrained and involves multiple objectives, so solving it once might not guarantee trajectory safety. So, we iteratively solve this problem using the re-guide strategy presented in Alg. \ref{reguide_algorithm} until the entire trajectory is collision-free. From our observations, the designed static collision and dynamic obstacle costs help the optimizer find the collision-free trajectory, but it might fail for the following reasons. First, the collision cost weights in Eqn. \ref{objective_function} are not large enough. For this scenario, Alg. \ref{reguide_algorithm} initializes the weights $\alpha_{\text{static}}$ and $\alpha_{\text{dynamic}} $ for static and dynamic collision cost with a user-defined cost inflate factor $\lambda$ (Lines \ref{cost_weights}-\ref{inflate_factor}) and increase the collision cost weights if they are too small (Lines \ref{inflate_static_weight} and \ref{inflate_dynamic_weight}).  Second, control points might be pushed towards new obstacles after optimization due to the "push force" from cost functions. In this case, we treat those control points as the re-guide required points as the previous collision cost and gradient approximations become invalid. So, Alg. \ref{reguide_algorithm} checks whether there exists the re-guide required points (Line \ref{check_reguide}) in each iteration and assigns new guide points to those control points (Line \ref{assign_new_guide_point}). The process will repeat until a collision-free trajectory is generated.

\section{Result and Discussion}
\begin{figure}[t] 
    \centering
    \includegraphics[scale=0.55]{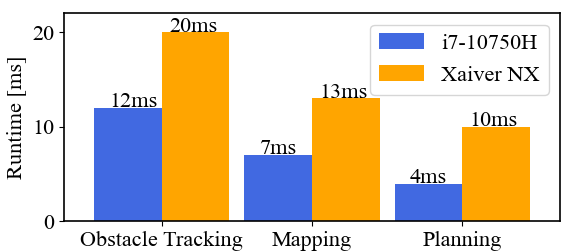}
    \caption{The recorded average runtime for each component of our system. The entire system is able to run in real-time by the onboard computer.}
    \label{runtime}
\end{figure}

To evaluate the proposed method's performance, we conduct simulation experiments and physical flight tests in dynamic environments. The algorithm implementation is based on C++ and ROS running on Intel i7-10750H@2.6GHz for simulation experiments and Nvidia Xavier NX for physical tests. The runtime for each component of our system is shown in Fig. \ref{runtime}. Overall, the entire system can run in real-time for the laptop and the UAV's onboard computer. The 3D dynamic map contains the obstacle tracking module and voxel mapping module, both using a small computational cost and taking less than $20$ms each iteration. The proposed planner can run up to around 100Hz by the UAV's onboard computer to guarantee a fast response to dynamic obstacles. 

\subsection{Obstacle Tracking Evaluation}
\begin{figure}[t] 
    \centering
    \includegraphics[scale=0.39]{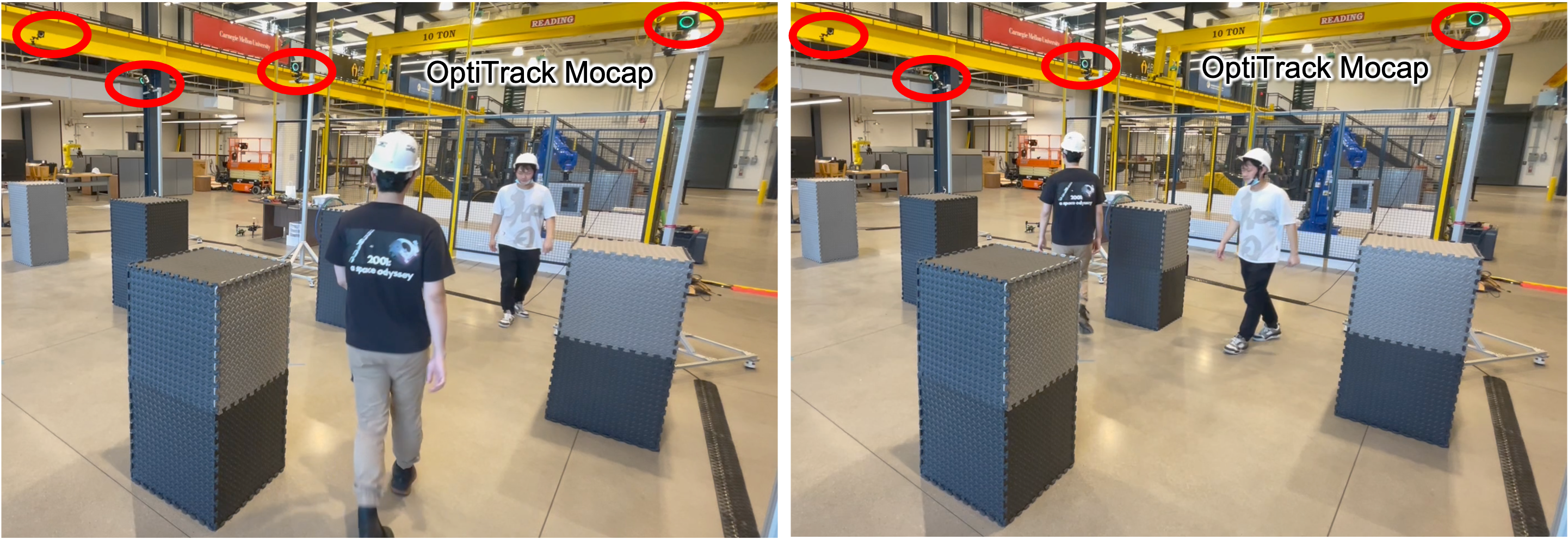}
    \caption{Illustration of using the OptiTrack motion capture system to estimate the error of the proposed dynamic obstacle detection and tracking method.}
    \label{mocap}
\end{figure}

To quantitatively evaluate the performance of our dynamic obstacle tracking method, we measure the errors in simulation and physical experiments using the external motion capture system (Fig. \ref{mocap}). The collected data is shown in Table \ref{detection_result}. The table results present the tracking algorithm's mean position, velocity, and obstacle size measurement errors. One can observe that both errors for position and velocity are relatively small in the simulation compared to the physical tests. This larger errors are mainly caused by the noises from physical depth images. The position and velocity errors in the real world are with the mean of 0.19$\text{m}$ and 0.21$\text{m/s}$, which are reasonable considering the low computation requirement and safe for dynamic obstacle avoidance. The obstacle size measurement errors are acceptable, with a mean of 0.25m for both simulation and physical tests. 

\begin{table}[h]
\begin{center}
\caption{Measurement of Detection and Tracking Errors.} \label{detection_result}
\begin{tabular}{ |c | c | c|  } 
 \hline

  Errors & Simulation Tests & Physical Tests \Tstrut\\ 
 \hline

 Position Error (\text{m})  & $0.09$  & $0.19$ \Tstrut\\ 
 \hline

 Velocity Error (\text{m}/\text{s}) & $0.10$  & $0.21$ \Tstrut\\  
 \hline
 
 Size Error (\text{m}) & $0.25$  & $0.25$ \Tstrut\\  
 \hline
\end{tabular}
\end{center}
\end{table}

\subsection{Simulation Experiments}
\begin{figure}[t] 
    \vspace{0.2cm}
    \centering
    \includegraphics[scale=0.0955]{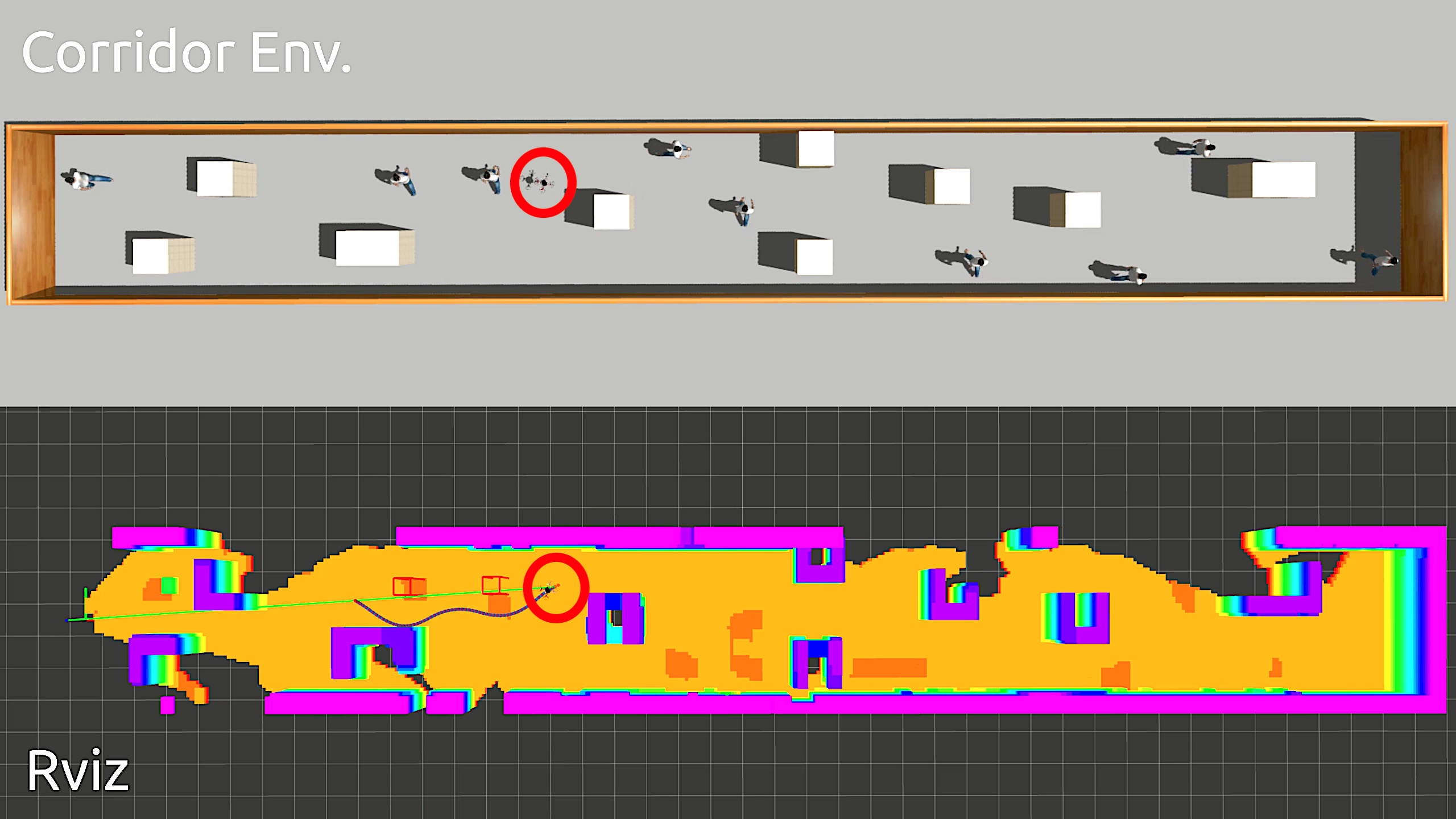}
    \caption{The corridor environment simulation. The robot (shown in the red circle) safely navigates through the environment without a pre-built map.}
    \label{corridor_simulation}
\end{figure}

The simulation experiments include three dynamic environments of various static structures. Fig. \ref{corridor_simulation} shows the corridor environment with the robot avoiding obstacles. We evaluate and benchmark the performance quantitatively using the collision, freezing, and success rates. The freezing rate denotes the rate of the robot failing to generate the trajectory and being forced to stop. For each environment, we select several target points for the robot to navigate and count the collision, freezing, and success times. We ran the experiments 20 times in the three environments (420 trajectories in total) and obtained the results in Table \ref{benchmark}. Note that we do not have the pre-built map for navigation, and the robot needs to navigate through the environments with only the given goal positions and observations from its sensor.

The state-of-the-art local planner \cite{egoplanner} is used for performance benchmarking. We also include our methods without vision detection (ViGO w/o Vision) and without the receding horizon distance field (ViGO w/o RH) to analyze the effects. From Table \ref{benchmark}, we can see that our proposed method has the highest total success rate. The EGO-Planner and our approach without vision detection have similar results in dynamic environments. From our observations, those failures are mainly from the latency of voxel map updates for moving obstacles, so adding vision-aided obstacle tracking to the system can largely improve the success rate. Besides, the performance of our method without the receding horizon distance field is slightly worse, with a lower success rate and higher collision rate. In the experiments, the method using the receding horizon distance field can make the robot act to obstacles earlier based on future predictions, which can help improve safety. In addition, our method has a higher freezing rate than its counterparts, which is the primary cause of failure. This phenomenon is reasonable since we apply the more strict constraints on dynamic obstacles, and it is effective as it increases the success rate. 

\begin{table}[h]
\begin{center}
\caption{The Benchmark of the collision rate, freezing rate, and success rate in the simulation experiments.} \label{benchmark}
\begin{tabular}{  c   c  c  c } 
 \hline

  Methods & Collision Rate & Freezing Rate & Success Rate \Tstrut\\ 
 
 \hline
 
 EGO-Planner \cite{egoplanner}  & $31.67\%$  & $2.14\%$ & $66.19\%$ \Tstrut\\ 
 ViGO w/o Vision & $34.52\%$ & $0.95\%$ & $64.53\%$ \\

 ViGO w/o RH & $7.14\%$  & $4.76\%$ & $88.10\%$\\

 \textbf{ViGO (Ours)} & $1.43\%$  & $5.95\%$ & $\textbf{92.62\%}$\\  

 \hline
\end{tabular}
\end{center}
\end{table}

\subsection{Physical Flight Tests}
\begin{figure}[t] 
    \centering
    \includegraphics[scale=0.51]{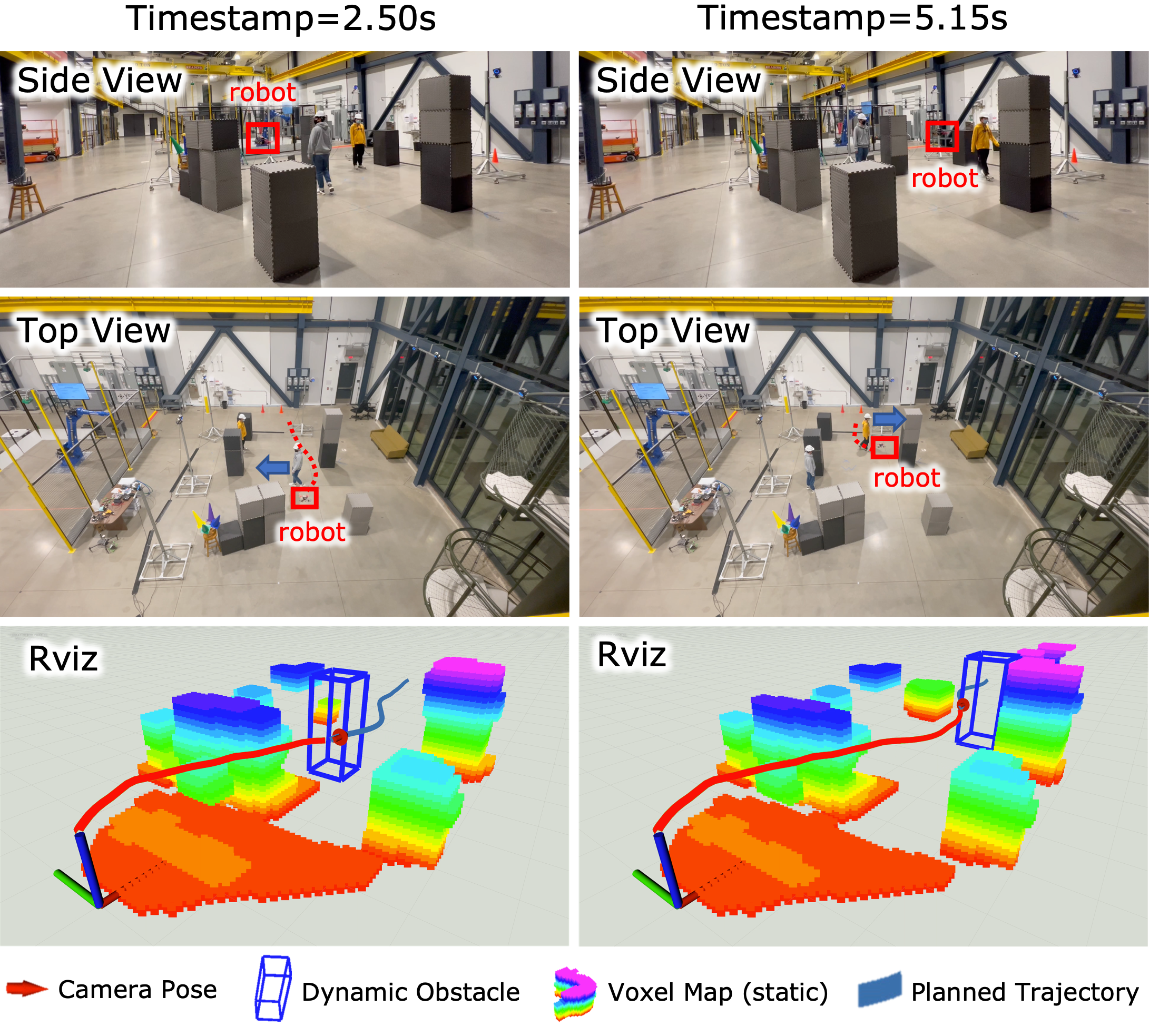}
    \caption{Illustration of a physical flight test in a dynamic environment. The side and top views from different timestamps are shown in the upper figures, and the Rviz visualization is presented at the bottom.}
    \label{real_flight}
\end{figure}

The physical experiments are conducted to verify the performance of the proposed method with the example target environments shown in Fig. \ref{intro_figure} and Fig. \ref{real_flight}. Our customized quadcopter (shown in Fig. \ref{intro_figure}) is equipped with an Intel Realsense D435i depth camera with Nvidia Xavier NX onboard computer. We apply the visual-inertial odometry (VIO) \cite{vins} to estimate the robot's position and velocity. The PX4-based flight controller is used to control the quadcopter to track the trajectories. The computation of all modules, including obstacle tracking, mapping, localization, and trajectory optimization, is performed onboard by the quadcopter.

\begin{figure}[t] 
    \centering
    \includegraphics[scale=0.59]{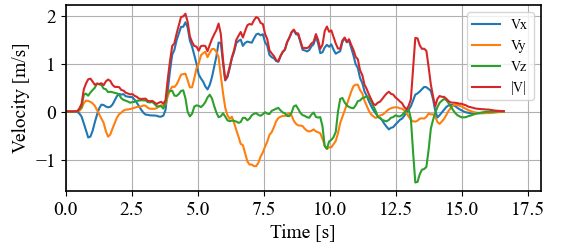}
    \caption{The velocity profile of a physical flight test for each axis. The velocity data are obtained using the onboard visual-inertial state estimation.}
    \label{velocity_plot}
\end{figure}

An example of a physical flight experiment is shown in Fig. \ref{real_flight}. In the experiment, we have two persons walking at around 1m/s and trying to block the robot's moving directions. The robot must navigate towards the goal location and avoid static and dynamic obstacles without a pre-built map. From the bottom two figures, we can see that the robot is able to detect and track the dynamic obstacles as the blue bounding boxes and build the static map shown as the colorful voxels. When the robot has potential collisions, it can successfully generate the trajectory (blue curves) to bypass static and dynamic obstacles. The robot history trajectory is presented as the red curves. The velocity profile of this physical flight test is shown in Fig. \ref{velocity_plot}. The data are collected using the onboard state estimation, and the maximum velocity is around 2m/s. By repeating the experiment with different static environment structures, we verify that our proposed methods can successfully perform navigation and obstacle avoidance in dynamic environments.

\section{Conclusion and Future Work}
This paper presents our vision-aided gradient-based B-spline optimization (\textbf{ViGO}) for navigating dynamic environments. The proposed method utilizes the vision-aided dynamic map to track dynamic obstacles. Our circle-based guide point algorithm is applied to estimate the costs and gradients for static collisions, while the receding horizon distance field is used to calculate dynamic obstacle costs. The proposed method follows the iterative re-guide optimization to generate a safe trajectory for navigation and obstacle avoidance. The simulation shows that our approach outperforms the state-of-the-art planner's success rate by using the vision obstacle tracking and receding horizon distance field. The 
physical experiments prove that our system can successfully avoid moving obstacles and run in real-time. For future improvement, we will focus on the 3D obstacle detection and tracking accuracy and apply multiple cameras to increase the tracking field of view. 
 
\section{Acknowledgement}
\noindent The authors would like to thank TOPRISE CO., LTD and Obayashi Corporation for their financial support in this work. 

\bibliographystyle{IEEEtran}
\bibliography{bibliography.bib}

\end{document}